\documentclass[10pt,twocolumn,letterpaper]{article}

\usepackage{iccv}
\usepackage{times}
\usepackage{graphicx}
\usepackage{amsmath}
\usepackage{amssymb}
\usepackage{booktabs}
\usepackage{makecell}
\usepackage{tabulary}

\usepackage{bbm}

\usepackage[dvipsnames]{xcolor,colortbl}
\usepackage{tikz}
\usepackage{comment}

\usepackage{color}
\usepackage{booktabs}
\usepackage{multirow}
\usepackage[accsupp]{axessibility}  %
\usepackage{caption}
\usepackage{subcaption}

\usepackage{dsfont}

\DeclareMathOperator*{\argmin}{arg\,min}
\usepackage[sort&compress,numbers,square,comma]{natbib}

\definecolor{Gray}{gray}{0.85}
\newcolumntype{a}{>{\columncolor{Gray}}c}

\usepackage[pagebackref=true,breaklinks=true,letterpaper=true,colorlinks,bookmarks=false]{hyperref}

\usepackage[capitalize]{cleveref}

\crefname{section}{Sec.}{Secs.}
\Crefname{section}{Section}{Sections}
\Crefname{table}{Table}{Tables}
\crefname{table}{Tab.}{Tabs.}

\iccvfinalcopy %

\def\iccvPaperID{9594} %

\ificcvfinal\fi

\begin{document}

\title{MOST: Multiple Object localization with Self-supervised Transformers for object discovery}

\author{Sai Saketh Rambhatla$\text{}^{1,3}$\footnotemark\\
{\tt\small rssaketh@meta.com}
\and
Ishan Misra$\text{}^{1}$\\
{\tt\small imisra@meta.com}
\and 
Rama Chellappa$\text{}^{2,3}$\\
{\tt\small rchella4@jhu.edu}
\and 
Abhinav Shrivastava$\text{}^{3}$\\
{\tt\small abhinav@cs.umd.edu}
\and
{$^{1}$Meta}\quad
{$^{2}$Johns Hopkins University}\quad
{$^{3}$University of Maryland, College Park}
}

\maketitle

\makeatletter
\def\blfootnote{\xdef\@thefnmark{}\@footnotetext}
\makeatother
\blfootnote{Work done while at UMD.}
\ificcvfinal\thispagestyle{empty}\fi

\begin{abstract}
We tackle the challenging task of unsupervised object localization in this work.
Recently, transformers trained with self-supervised learning have been shown to exhibit object localization properties without being trained for this task. In this work, we present \textbf{M}ultiple \textbf{O}bject localization with \textbf{S}elf-supervised \textbf{T}ransformers (MOST) that uses features of transformers trained using self-supervised learning to localize multiple objects in real world images. MOST analyzes the similarity maps of the features using box counting; a fractal analysis tool to identify tokens lying on foreground patches. The identified tokens are then clustered together, and tokens of each cluster are used to generate bounding boxes on foreground regions.
Unlike recent state-of-the-art object localization methods, MOST can localize multiple objects per image and outperforms SOTA algorithms on several object localization and discovery benchmarks on PASCAL-VOC 07, 12 and COCO20k datasets. Additionally, we show that MOST can be used for self-supervised pre-training of object detectors, and yields consistent improvements on fully, semi-supervised object detection and unsupervised region proposal generation.Our project is publicly available at this \href{https://rssaketh.github.io/most}{website}.
\end{abstract}

\section{Introduction}
Object detectors are important components of several computer vision systems such as visual relationship detection~\cite{lu2016visual, Krishna2016VisualGC}, human-object interaction detection~\cite{bansal2019detecting, Tamura2021QPICQP, Wang2020LearningHI, Gkioxari2018DetectingAR}, scene graph generation~\cite{Xu2017SceneGG} and object tracking~\cite{Wojke2017simple, wang2019towards} etc. Performance of object detectors is heavily reliant on the availability of training data. However, annotating large object detection datasets can be expensive and time consuming~\cite{lin2014coco, gupta2019lvis}.
Additionally, the vocabulary of object detectors is limited by the training datasets and such detectors  often fail to generalize to new categories~\cite{Dhamija_2020_WACV}.
This strategy is not scalable and a more effective approach is warranted. Object discovery is one such task that has the potential to address these concerns.
\begin{figure}[t]
    \centering
    \includegraphics[width=\linewidth]{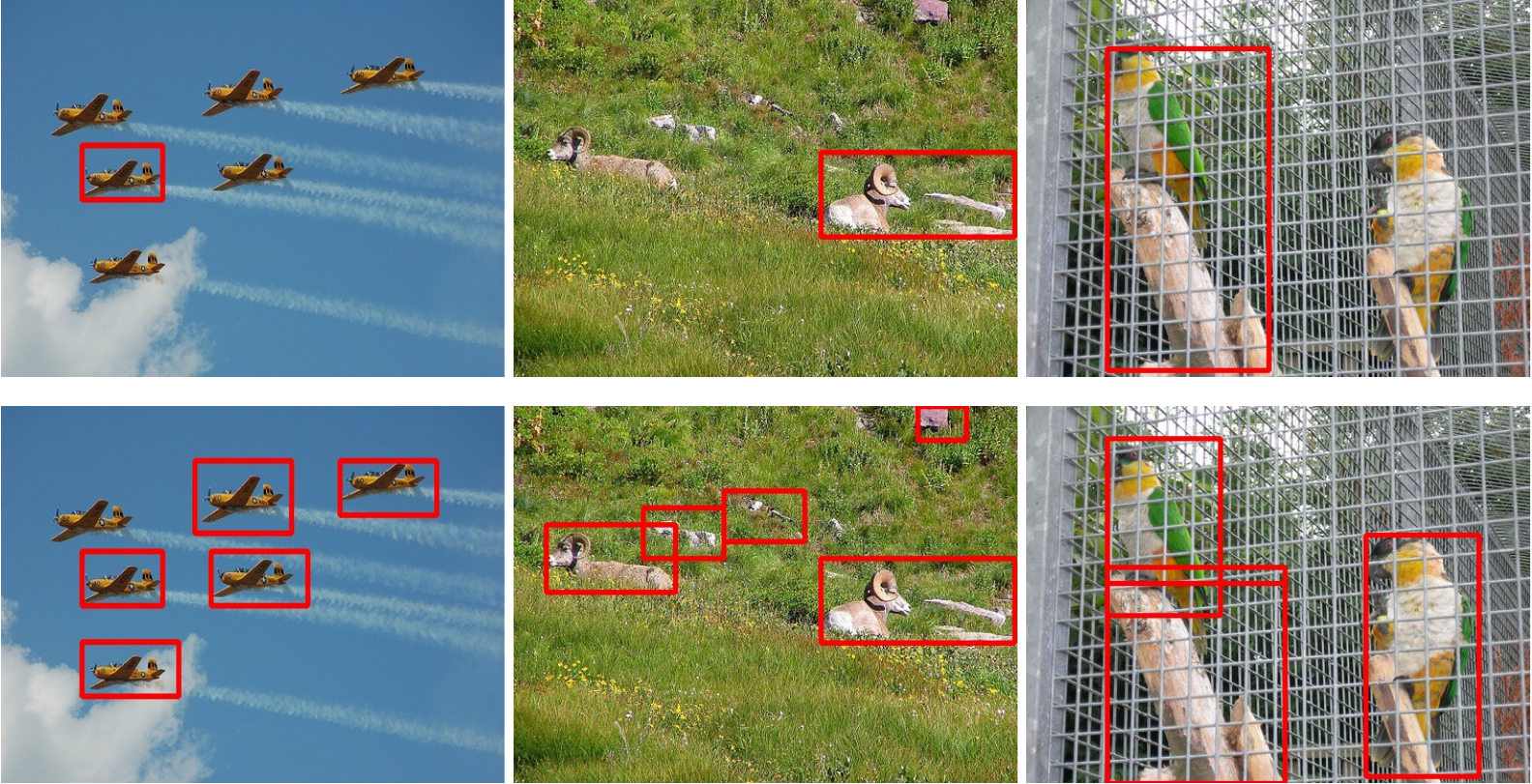}
    \caption{\textbf{Top}: Methods like LOST~\cite{LOST} (shown in figure), TokenCut~\cite{wang2022tokencut} identify and localize the most salient foreground object and hence can detect only one object per image. \textbf{Bottom}: MOST is a simple, yet effective method that localizes multiple objects per image without training. }
    \label{fig:teaser}\vspace{-1.5em}
\end{figure}
Object discovery is the problem of identifying and grouping objects/parts in a large collection of images without human intervention~\cite{LeeG11, Romea-2014-17150, 5540237, 1541280}. The first step in object discovery is to obtain region proposals and subsequently group them semantically. Previous works on discovery used Selective Search~\cite{Uijlings13}, randomized Prim's~\cite{6751426} or a region proposal network (RPN)~\cite{NIPS2015_5638} to get object proposals.  \cite{Vo20rOSD, VoBCHLPP19, Vo21LOD} used inter-image similarity to construct a graph and performed optimization or ranking, to localize objects without any supervision. Such methods are computationally expensive and often fail to scale to datasets larger than 20000 images. \cite{Rambhatla2021ThePO} used region proposals from an RPN and proposed a never ending learning approach and is the first method shown to scale to $\sim$100000 images. However, these region proposal methods are often of low quality, and therefore reduce the performance of discovery systems. Recently, LOST~\cite{LOST} and TokenCut~\cite{wang2022tokencut} leveraged the object segmentation properties of transformers~\cite{attisall} trained using self-supervised learning (DINO~\cite{caron2021emerging}) to obtain high quality object proposals. They demonstrate significant improvements over state-of-the-art on object discovery, salient object detection and weakly supervised object localization benchmarks.

However, both LOST~\cite{LOST} and TokenCut~\cite{wang2022tokencut} assume the presence of a single salient object per image and hence, can localize only one object as shown in Fig \ref{fig:teaser} (top). This assumption may hold for object centric datasets like ImageNet~\cite{ILSVRC15} but is not true for scene-centric real world datasets like PASCAL-VOC~\cite{cite-key} and COCO~\cite{lin2014coco}. In this work, we address the problem of localizing multiple objects per image and demonstrate the effectiveness of our approach for the task of unsupervised object localization and discovery on several standard benchmarks. 

We propose a new object localization method called ``Multiple Object localization with Self-supervised Transformers'' (MOST) which is capable of localizing multiple objects per image without using any labels. We use the features extracted from a transformer~\cite{attisall} network trained with DINO~\cite{caron2021emerging}. Our method is based on two empirical observations; 1) Patches within foreground objects have higher correlation with each other than the ones on the background~\cite{LOST} and 2) The similarity map computed using the features of a foreground object with all the features in the image is usually more localized and less noisier than the one computed using the feature of a background. Our algorithm analyzes the similarities between patches exhaustively using a fractal analysis tool called box counting~\cite{Liu2003FractalDI}. This analysis picks a set of patches that most likely lie on foreground objects. Next, we perform clustering on the patch locations to group patches belonging to a foreground object together. Each of these clusters is called \textit{pools}. A binary mask is then computed for each \textit{pool} and a bounding box is extracted. This capability enables the algorithm to extract multiple bounding boxes per image as shown in Fig.\ref{fig:teaser} (bottom). We demonstrate that \textbf{without any training}, our method can outperform state-of-the-art object localization methods that train class agnostic detectors to detect multiple objects. 
To prove the effectiveness of MOST, we demonstrate results on several object localization and discovery benchmarks. On self-supervised pre-training for object detectors, using MOST yields consistent improvement across multiple downstream tasks using $6\times$ fewer boxes. When compared against other self-supervised transformer-based localization methods, MOST achieves higher recall  with and without additional training.
We summarize the contributions of our work below.

\begin{itemize}
    \item We propose MOST, an effective method to localize and discover multiple objects per image without supervision using transformers trained with DINO. 
    \item We perform exhaustive experiments to assess the performance of MOST on several localization and discovery benchmarks and show significant improvements over the baselines. 
\end{itemize}
The paper is organized as follows. In Section~\ref{sec:related} we discuss related works on object localization and discovery. We describe our approach in detail in Section~\ref{sec:method}. We describe our experimental setup and present results in Section~\ref{sec:experiments} and conclude in Section~\ref{sec:conclusion}.

\section{Related Works}\label{sec:related}
\noindent\textbf{\underline{Unsupervised Object Localization and Discovery}}:
Object category discovery can be broadly segregated into image-based~\cite{HsuLK16,HsuLK18,HsuLSOK19,han19learning,han20automatically,singh-cvpr2019} and region-based methods~\cite{VoBCHLPP19,Vo20rOSD,Vo21LOD,LeeG11,5540237,hal-01110036,hal-01153017,LOST,wang2022tokencut,doersch2014context,Rambhatla2021ThePO}. Region-based methods start by generating object proposals and later group them into coherent semantic groups. Image-based approaches on the other hand, assume the localization task to be solved and focus solely on the semantic grouping. Our current method is closer to the former. Vo \etal,~\cite{VoBCHLPP19,Vo20rOSD,Vo21LOD} localize regions in images by constructing an inter-image similarity graph between regions and partitioning it using optimization or ranking. Due to the quadratic nature of the graph, these methods cannot scale to large datasets beyond tens of thousands of images. Our current work does not compute inter-image similarity between regions and scales linearly with number of images. Lee \etal,~\cite{LeeG11} propose object graphs that incorporates knowledge about a few known categories to facilitate the discovery of novel categories. MOST doesn't assume any prior knowledge and has the ability to discover objects from scratch. Lee \etal,~\cite{5540237} extend object graphs to a curriculum based discovery pipeline, that learns to discover easy objects first and progressively proceeds to discover the harder ones. Along similar lines, Rambhatla \etal,~\cite{Rambhatla2021ThePO} propose a large scale discovery pipeline, similar to ~\cite{LeeG11} that leverages prior knowledge about a few categories. Authors of~\cite{Rambhatla2021ThePO} use an RPN~\cite{NIPS2015_5638} trained on known categories as the localization method. In contrast to that, MOST localizes objects in images using features of a transformer~\cite{attisall} trained with DINO~\cite{caron2021emerging}.

\noindent\textbf{\underline{Object localization using self-supervised networks}}:
Recently, CNNs~\cite{He2016DeepRL} and Transformers~\cite{attisall} trained in a self-supervised fashion, have been shown to exhibit object localization/segmentation properties~\cite{caron2021emerging,Dwibedi_2021_ICCV}. This property is of particular interest as it has the potential to facilitate research on unsupervised localization, detection and segmentation. \cite{LOST} is a simple method, based on the observation that the key features of the last self attention layer of a transformer, trained using DINO, has certain affinity. They construct a map of inverse degree to extract bounding boxes on objects in an unsupervised fashion. This method is shown to outperform recent state-of-the-art methods by a significant margin. \cite{wang2022tokencut} propose an alternate method for localizing objects using self-supervised transformers, based on normalized cut~\cite{Shi1997NormalizedCA}. 
TokenCut~\cite{wang2022tokencut} construct an undirected graph based on token feature similarities and use normalized cut to cluster foreground and background patches. Spectral clustering is used to solve the graph-cut and they show that the eigen vector corresponding to the second smallest eigenvalue provides a good cutting solution. TokenCut outperforms LOST on unsupervised object discovery. In addition to discovery,~\cite{wang2022tokencut} also demonstrate impressive results on unsupervised saliency detection and weakly supervised object localization. Kyriazi \etal,~\cite{melas-kyriazi22deep} propose deep spectral methods, that performs normalized cut~\cite{Shi1997NormalizedCA} but using an affinity matrix computed using semantic and color features. Since this method is very similar to TokenCut and achieves lower performance, we only compare with the latter in this work.

However, one limitation of LOST and TokenCut is that they can localize only one object per image. Our proposed method, MOST can automatically localize multiple objects per image and outperforms LOST and TokenCut on standard discovery benchmarks.

\begin{figure}[t]
    \centering
    \includegraphics[width=\linewidth]{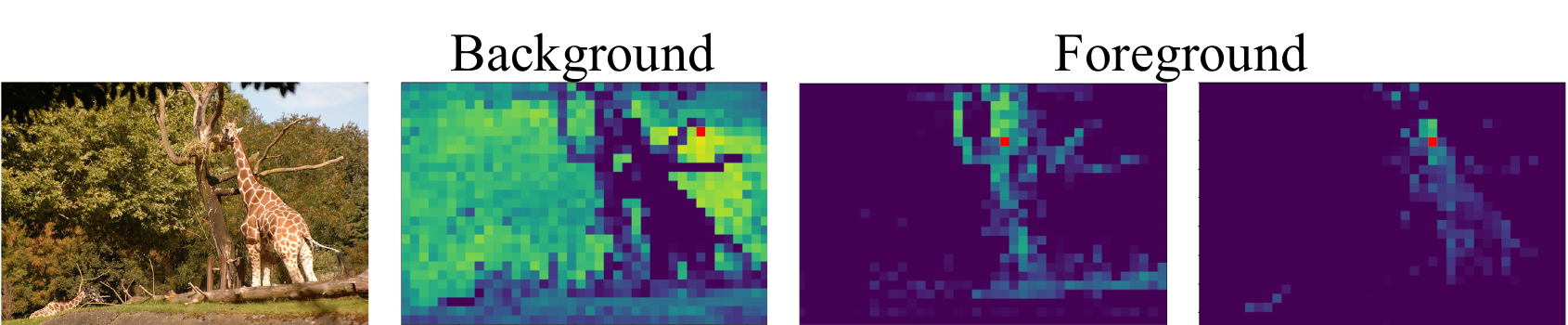}
    \caption{\textbf{Motivation for MOST}: Example showing similarity maps of tokens within background and foreground for an image from the COCO dataset. Similarity maps of tokens within foreground patches are less random spatially.}
    \label{fig:motiv}
\end{figure}
\section{Approach: MOST}\label{sec:method}
\begin{figure*}[t]
    \centering
    \includegraphics[width=\linewidth]{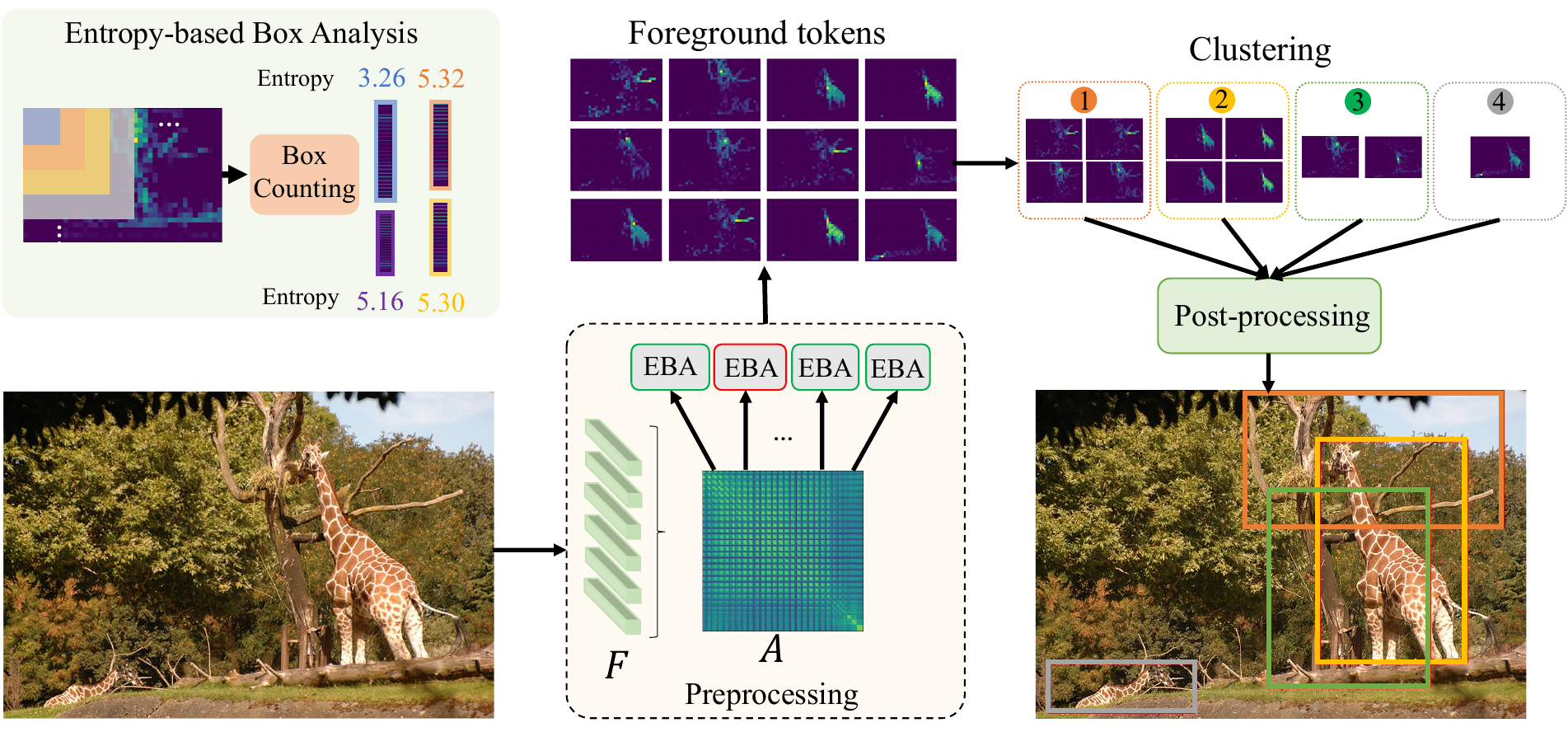}
    \caption{\textbf{Overview of MOST}: MOST operates on features extracted from transformers trained using DINO. The features are used to compute the outer product $A$. Each row of $A$ is analyzed by the entropy-based box analysis (EBA) module that identifies tokens extracted from foreground patches. These patches are clustered using spatial locations as features to form \textit{pools}. Each \textit{pool} is then post-processed to generate a bounding box. }
    \label{fig:pipeline}
\end{figure*}
MOST can be used to localize multiple objects in an image. Our approach, illustrated in Fig. \ref{fig:pipeline}, first identifies a set of tokens that is computed from patches on foreground objects. These tokens are then clustered and each cluster, named \textit{pool}, is used to obtain a bounding box. Next, we discuss a few preliminaries in Section~\ref{subsec:prelim} followed by the motivation and proposed solution in Section~\ref{subsec:approach}.
\subsection{Preliminaries}\label{subsec:prelim}

\noindent\textbf{\underline{Box Counting}}: Box counting is a method of analyzing a pattern by breaking and analyzing it at smaller scales. This is done by performing a raster scan of the pattern at different scales and computing appropriate metrics to assess its fractal nature. In this work, we use a sliding window scan.

\noindent\textbf{\underline{Vision Transformers}}: ViTs~\cite{dosovitskiy2020vit} operate on learned embeddings, called tokens, generated from non-overlapping image patches of resolution P$\times$P (typically P=8/16) that form a sequence. To be precise, an image I of shape $H\times W\times3$ is first divided into non-overlapping patches of resolution $P\times P\times$3, generating a total of N = $HW/P^{2}$ patches. Next, each patch is embedded via a trainable linear layer to generate a token of dimension $d$ to form a sequence of patches. An extra $[\texttt{CLS}]$ token~\cite{devlin-etal-2019-bert} is added to this sequence, whose purpose is to aggregate the information from the tokens of the sequence. Typically, a projection head is attached to the $[\texttt{CLS}]$ to train the network for classification. 

\noindent\textbf{\underline{DINO}}: DINO~\cite{caron2021emerging} combines self-training and knowledge distillation without labels for self supervised learning. DINO constructs two global views and several local views of lower resolution, from an image. DINO consists of a teacher and a student network. The student processes all the crops while the teacher is operated only on the global crops. The teacher network then distills its dark knowledge to the student. This encourages the student network to learn local to global correspondences. In contrast to other knowledge distillation methods, DINO's teacher network is updated dynamically during training using exponential moving average. DINO was shown to perform on par or better than several baselines on several tasks of image retrieval, copy detection, instance segmentation etc. The property of importance to the current work, is the ability of DINO to localize foreground regions of semantic entities in an image. \cite{LOST, wang2022tokencut} leverage this property to localize the salient object in an image by using the key features from the last self-attention layer. Similar to~\cite{LOST, wang2022tokencut}, we concatenate the key features of all the heads in the last self-attention layer to obtain the input to MOST.

\subsection{Multiple object localization}\label{subsec:approach}

\noindent\textbf{\underline{Motivation}}: 
Consider the example shown in Fig.~\ref{fig:motiv}. We show three examples of the similarity maps of a token (shown in red) picked on the background (column 2) and foreground (columns 3, 4). Tokens within foreground patches have higher correlation than the ones on background~\cite{LOST}. This results in the similarity maps of foreground patches being \textit{less} ``spatially" random than the ones on the background. The task then becomes to analyze the similarity maps and identify the ones with less spatial randomness. Box counting~\cite{plotnick1996lacunarity, Li2009AnIB} is a popular technique in fractal analysis that analyzes spatial patterns at different scales to extract desired properties. Hence, we adopt box counting for our case and since, we are interested in randomness, we adopt entropy as the metric.

\noindent\textbf{\underline{Preprocessing}}: The input to our method is a $d$-dimensional feature $F\in \mathbb{R}^{N\times d}$, extracted from an image using a neural network. Here, $N$ denotes the number of spatial locations in the feature map, in case of a CNN, or number of tokens, in case of a transformer network. The aim is to identify subsets of tokens, which we call \textit{pools}, that can be used to localize all the objects in an image. We do not make any assumption on the number of objects present in the image. 
Given the feature $F$, we compute an outer product matrix $A = FF^T \in\mathbb{R}^{N\times N}$. Row $i$ of matrix $A$, i.e., $A[i,:]$ encodes a similarity map of a token at location $i$ with all the other tokens in $F$. Next, each row of $A$ is processed by the Entropy-based Box Analysis (EBA) module.

\noindent\textbf{\underline{Entropy-based Box Analysis (EBA)}}: The proposed entropy based box analysis module performs a fractal analysis method, called box counting to segregate similarity maps of tokens on foreground patches from those of background. As shown in Fig. \ref{fig:pipeline}, we perform a raster scan with increasing box (used interchangeably with kernel in this work) sizes on each map. Traditionally, measures like lacunarity~\cite{smith_t_g_1996_1259855} are computed within each box to analyze the pattern. In this work, we average the elements within each box. This can be implemented efficiently using pooling operations. The resulting downsampled map is flattened and the entropy is computed using the pmf computed as follows: $p(x = x_i) = \Sigma_{i=1}^{h.w} \frac{\mathbbm{1}(f_i == x_i)}{h.w}$, where $f_i$ is the $i^{\text{th}}$ index in the feature map. A downsampled map belongs to a token on a foreground patch if its entropy is less than a threshold $\tau$. Using $K$ boxes in the EBA module results in $K$ entropy values $e_k (k\in\{1,2,\cdots,K\})$. Finally, we perform a majority voting among the entropies of all the downsampled maps, \ie, $\Sigma_{i=1}^{K} \frac{\mathbbm{1}(e_i \leq \tau)}{K} > 0.5$, to decide if the original similarity map belongs to a token on a foreground patch. A map of dimension $n\times n$ has a maximum entropy of $log(n^2)$. We use a threshold of the form $\tau=a+blog(n^2)$ (we use $a=1, b=0.5$ in this work). We do not consider $\tau$ as a hyperparameter and we pick a value that is mid-way between the minimum and maximum permissible value (b=0.5). To prevent a threshold of $0$ for $n=1$, we add a constant (a=1). 

\noindent\textbf{\underline{Clustering}}: The EBA module, identifies a set $\mathcal{S} = \{p | p\in\{1,2,\cdots,N\}\}$, that contains the spatial locations of tokens computed from foreground patches. Often, highly redundant neighboring tokens are identified. We group neighboring tokens with the help of a clustering step to obtain \textit{pools}. We convert the linear index $p$ of the tokens to cartesian coordinates $(x,y)$, and use that as the feature for clustering. Manhattan distance is used as the dissimilarity metric with a threshold $\epsilon$ ($\epsilon=2$ i.e. Moore neighborhood). Since, the number of pools is not known a-priori, we use a density-based clustering method, DBSCAN~\cite{10.5555/3001460.3001507} which automatically identifies the number of clusters from the data. \textit{Pools} identified by the clustering step are then post-processed to obtain bounding boxes on foreground objects.
\begin{table*}[!ht]

\renewcommand{\arraystretch}{1.2}
\setlength{\tabcolsep}{8pt}
\centering
\caption{\textbf{Results on unsupervised pre-training of object detectors}. We train object detectors in a self-supervised fashion on the COCO dataset using different localization methods and compare their performance on the downstream tasks of semi and fully supervised object detection. COCO train set is used for fine-tuning and $k\%$ refers to the number of labeled samples used for training. Results are reported using AP[0.50:0.95] (denoted as AP), AP$_{0.50}$, and AP$_{0.75}$ on COCO validation set.} \label{tab:od-1}
\resizebox{\linewidth}{!}{
\begin{tabular}{@{}lccccccccccccc@{}}
\toprule
\multirow{3}{*}{Method} & \multirow{3}{*}{\makecell{Boxes \\ per \\ image}} &\multicolumn{6}{c}{\textbf{VOC 07+12 }} &\multicolumn{6}{c}{\textbf{COCO}}  \\
\cmidrule[\cmidrulewidth](l){3-8}
\cmidrule[\cmidrulewidth](l){9-14}
&  & \multicolumn{3}{c}{$k=10\%$} & \multicolumn{3}{c}{fully supervised} & $k=1\%$ & $k=2\%$ & $k=5\%$ & \multicolumn{3}{c}{fully supervised} \\
  \cmidrule[\cmidrulewidth](l){3-5}
  \cmidrule[\cmidrulewidth](l){6-8}
 \cmidrule[\cmidrulewidth](l){9-11}
 \cmidrule[\cmidrulewidth](l){12-14}
&  & AP & AP$_{50}$ & AP$_{75}$ & AP & AP$_{50}$ & AP$_{75}$ & \multicolumn{3}{c}{AP} & AP & AP$_{50}$ & AP$_{75}$  \\
 \midrule
LOST~\cite{LOST} & 1 & 40.88	& 60.31 & 	44.36& 63.58 & 83.27 & 70.48 & 12.83 $\pm$ 0.32&	17.23 $\pm$ 0.30&	23.43 $\pm$ 0.38&44.30 & 62.80 & 48.40 \\
TCut~\cite{wang2022tokencut} & 1 &  41.14  & 60.59& 44.35	& 63.79 & 83.56 &70.70 	&13.13 $\pm$ 0.38& 17.27	$\pm$0.21 &	23.27	$\pm$ 0.23& 43.80	&62.30 & 47.50\\
\multirow{3}{*}{SS~\cite{Uijlings13}} & 5 & 39.12  & 57.51 & 42.29 & 63.44 & 83.14 & 70.35 &13.57 $\pm$ 0.38&	17.87 $\pm$ 0.32	&23.17 $\pm$ 0.40 &44.30 & 62.80 & 48.10 \\
 & 10 & 40.76  & 60.00 & 44.46  & 64.23 & 83.44 & 71.55  & 13.73 $\pm$ 0.29&	18.00 $\pm$	0.26	&22.83 $\pm$ 0.25&43.90 & 62.60 & 47.60\\
 & 15 & 42.14 & 61.41 &	45.86	&64.24	&83.74	&71.41 & 13.87 $\pm$ 0.29 &	18.23 $\pm$ 0.40	&23.13 $\pm$	0.11&44.30&62.60	&48.30\\
MOST & 4.65 & 43.03 & 63.29	&46.61	& 64.34 &	84.12	& 71.77 &13.93 $\pm$ 0.38&	18.13 $\pm$	0.25	&22.63 $\pm$	0.11&44.80& 63.50	& 49.00 \\
\hline

SS & 30 & 42.12 & 61.20	&45.71 &	64.84&	83.98&	71.76&14.47 $\pm$ 0.35&	18.23 $\pm$	0.42	&\textbf{23.57 $\pm$	0.21} &44.00 &	62.30&	47.80\\
MOST & 13.09 & \textbf{44.40} & \textbf{63.83}	&\textbf{48.28} &	\textbf{65.24}&	\textbf{84.24}&	\textbf{72.37} &\textbf{14.83 $\pm$ 0.21}&	\textbf{18.30 $\pm$	0.17} &23.43 $\pm$ 0.45& \textbf{45.20} &	\textbf{64.00}	&\textbf{49.00}\\
\bottomrule
\end{tabular}
}

\end{table*}

\noindent\textbf{\underline{Post-processing}}: The clusters, called \textit{pools}, obtained from the clustering step are then post-processed to obtain one box per \textit{pool}. Consider $M$ \textit{pools} identified by the clustering step, i.e. $\mathcal{C}_i$, where $i\in\{1,2,\cdots M\}$. Each \textit{pool} $\mathcal{C}_i$ is a set of token locations $\mathcal{C}_i = \{p^i | p^i \in\{1,2,\cdots,N\}\}$.
We leverage the first observation mentioned above to obtain a bounding box from the \textit{pool} as follows.
First, we build a binary similarity matrix $\hat{A} = A > 0$. Next, within the tokens in the pool, we identify the one with lowest degree in $\hat{A}$, called the \textit{core} token, $c^*$. 
$$c^* = \argmin_{c\in\mathcal{C}_i} d_c \quad \text{where} \quad d_c = \sum_{j=1}^N \hat{A}[c,j]$$
Authors of LOST~\cite{LOST} report that tokens with low degrees most likely fall within an object. Next, we remove the tokens from the pool that do not correlate positively with $c^*$ to form a reduced \textit{pool} $\mathcal{C}^*_i$. This ensures that all the tokens in the current pool lie on the same foreground object.
Next, a binary mask is constructed by computing the sum of similarities of token features in $\mathcal{C}^*_i$ with the features of all the tokens, i.e. $m^i_k = \mathds{1}(\sum_{c\in\mathcal{C}^*_i} f_k^Tf_c \geq 0)$. Finally, connected component analysis is performed on the binary mask and the bounding box of the island that contains $c^*$ is selected as the region containing the object.
We repeat this process for all the $M$ \textit{pools} to generate $M$ bounding boxes per image. Note that, $M$ is not assumed to be known a-priori and is decided automatically by our method.
Additionally, we remove trivial boxes i.e., boxes which have area less than than a threshold (256) or cover the whole image.

\noindent\textbf{\underline{Implementation Details}}:
For all our experiments, we use the ViT-S/16 and ViT-B/8~\cite{dosovitskiy2020vit} models trained with DINO~\cite{caron2021emerging} to extract the features. We concatenate the key features of all the heads from the last self-attention layer to use as the input to our method.

\section{Experiments}\label{sec:experiments}

In this section we describe, in detail, the experimental setup used for evaluation. We evaluate our method on two setups, namely the localization setup, and the discovery setup. We begin by describing the datasets and metrics in Sec.~\ref{subsec:data}. We describe the evaluation setups in Sec.~\ref{subsec:setup}. Sec.~\ref{subsec:results} compares our method against contemporary work. We then describe ablation experiments in Sec.~\ref{subsec:ablation} and show qualitative results in Sec.~\ref{subsec:q_res}.

\begin{table*}[!t]
\begin{minipage}[b]{0.3\linewidth}
\renewcommand{\arraystretch}{1.2}
\setlength{\tabcolsep}{2pt}
\centering
\footnotesize
\caption{\textbf{Unsupervised class agnostic region proposal evaluation on COCO validation set}: We compare the performance of region proposals for training DETReg. R$k$ is Recall$@k$}\label{tab:caop}
\resizebox{\linewidth}{!}{
\begin{tabular}{@{}lccccccc@{}}
\toprule
Method & \makecell{Boxes per\\image} & AP&AP$_{50}$&AP$_{75}$ & R1 & R10 &R100 \\
 \midrule
LOST~\cite{LOST} & 1 & 0.1 &	0.5 &	0 &	0.4 &1.4 &3.9\\
TCut~\cite{wang2022tokencut} & 1& 0.3&	1&	0.1	&\textbf{0.6}&	\textbf{1.9}&	\textbf{4.6} \\
\multirow{3}{*}{SS~\cite{Uijlings13}} & 5& 0.1&	0.4&	0&	0.1&	1&	4.2 \\
 & 10 & 0.1&	0.3&	0	&0.1&	1.1&	4.4\\
 & 15 & 0.1&	0.3&	0	&0.1&	1	&4.1 \\
 & 30 & 0.1&	0.3&	0&	0.1&	1&	4\\
MOST& 4.65 & \textbf{0.8}& 	\textbf{1.4}	& \textbf{1}& 	\textbf{0.6}& 	\textbf{1.9}	& 4.4\\
\bottomrule
\end{tabular}
}
\end{minipage}\quad
\begin{minipage}[b]{0.39\linewidth}
\renewcommand{\arraystretch}{1.5}
\setlength{\tabcolsep}{1pt}
\centering
\footnotesize
\caption{\textbf{Results on object discovery}: Comparison of MOST with recent works on unsupervised object discovery. We experiment with three cluster numbers, \ie, 20, 30, 40, on VOC 2007, 2007+12 and 80, 90, 100 on COCO20k.}\label{tab:od1}
\resizebox{\linewidth}{!}{
\begin{tabular}{@{}llccccccccc@{}}
\toprule
 Metric & Train \textrightarrow& \multicolumn{3}{c}{\textbf{VOC 2007}} & \multicolumn{3}{c}{\textbf{VOC 07+12}}& \multicolumn{3}{c}{\textbf{COCO20k}}  \\
  \cmidrule[\cmidrulewidth](l){3-5}
 \cmidrule[\cmidrulewidth](l){6-8}
 \cmidrule[\cmidrulewidth](l){9-11}
  &Clusters \textrightarrow &20  & 30 & 40 & 20 &  30 & 40 & 80 & 90 & 100  \\
 \midrule
\multirow{2}{*}{\textcolor{gray}{AP}}& LOST~\cite{LOST} &9.15  &9.64& 10.11& \textbf{10.95}  & 12.14 & 12.97 & 2.66& 2.91& 2.86 \\
 &MOST & \textbf{9.20} & \textbf{10.07} & \textbf{11.09}& 10.12  & \textbf{12.89} & \textbf{13.30} & \textbf{3.13} & \textbf{3.18} &\textbf{3.32} \\
\midrule
\multirow{2}{*}{\textcolor{gray}{AP$_{50}$}} & LOST~\cite{LOST}&\textbf{26.32} &27.78& 29.46& \textbf{29.35} &33.27 &\textbf{34.80} &7.17 &7.72 &7.87\\
 
&MOST & 25.35& \textbf{28.19}& \textbf{31.31} & 27.04 &\textbf{34.40}&34.54&\textbf{ 8.13} & \textbf{8.14} &\textbf{ 8.76}\\
\bottomrule
\end{tabular}
}
\end{minipage}\quad
\begin{minipage}[b]{0.28\linewidth}
\renewcommand{\arraystretch}{1.0}
\setlength{\tabcolsep}{2pt}
\centering
\caption{\textbf{Results on single-object localization}: Comparison of MOST with recent object discovery methods on VOC 07, 12 and COCO20k using CorLoc. }\label{tab:corloc}
\resizebox{\linewidth}{!}{
\begin{tabular}{@{}lccc@{}}
\toprule
 \textbf{Method}& \textbf{VOC 07} & \textbf{VOC 12}& \textbf{COCO20k}  \\
 \midrule
 rOSD~\cite{Vo20rOSD} & 54.5 & 55.3 & 48.5  \\
 LOD~\cite{Vo21LOD} & 53.6 & 55.1 & 48.5 \\
 DINO-seg$^\dagger$~\cite{caron2021emerging} & 45.8 & 46.2 & 42.1\\
LOST~\cite{LOST} & 61.9 & 64.0 & 50.7\\
TokenCut~\cite{wang2022tokencut} & 68.8& 72.1 & 58.8 \\
\midrule
 LOST~\cite{LOST} + CAD & 65.7 & 70.4 & 57.5\\
 TCut~\cite{wang2022tokencut}+CAD & 71.4& 75.3 &62.6 \\
 \midrule
 MOST & \textbf{74.8} & \textbf{77.4} & \textbf{67.1}\\
\bottomrule
\end{tabular}
}
\end{minipage}
\end{table*}

\subsection{Datasets and Metrics}\label{subsec:data}
We use the PASCAL-VOC~\cite{cite-key} (2007, 2012 splits) and the COCO~\cite{lin2014coco} (COCO20k~\cite{Vo20rOSD} and COCO splits) datasets in our experiments. The PASCAL VOC~\cite{cite-key} 2007 and 2012 trainval sets consists of 5011, 11540 images respectively, spanning twenty objects. The PASCAL VOC~\cite{cite-key} test set consists of 4952 images. 
The COCO~\cite{lin2014coco} 2014 train set consists of ${\sim}$110k images containing over eighty objects and the COCO minival set consists of 5000 images.
We do not use any class or bounding box annotations for our method except for evaluation.

For the \textit{localization} setup, we use the average precision at different thresholds ([0.5:0.95], 0.5 and 0.75), average recall (AR\@1, AR\@10 and AR\@100) and Correct Localization (CorLoc) metrics for evaluation. CorLoc is defined as the fraction of the images in which atleast one object is localized with an IoU greater than a threshold (0.5 in this work). AP, AR are defined in the usual way. 
For the object discovery setup, we report both the PASCAL VOC style $\text{AP}_{50}$ and COCO style $\text{AP}_{[50:95]}$ metrics along with area under the purity-coverage plots~\cite{doersch2014context, Rambhatla2021ThePO}. 
We refer the interested readers to~\cite{Rambhatla2021ThePO} for definitions of purity and coverage.

\subsection{Setups}\label{subsec:setup}

\noindent\textbf{\underline{Localization setup}}:
This setup evaluates the localization performance of methods. We evaluate models on a) unsupervised pre-training, b) Multiple Object Localization, and c) single object localization. For unsupervised pre-training, localization methods are used to train object detectors in an unsupervised fashion and their performance is evaluated on the downstream task of object detection. In this work, we use the recently proposed DETReg~\cite{bar2021detreg} as the pre-training strategy which uses a Deformable DeTR~\cite{zhu2021deformable} architecture. DETReg uses an object localization method and pre-trains an object detector in an unsupervised fashion. We evaluate the pre-trained model on the downstream tasks of semi-supervised, fully-supervised and class-agnostic object proposal generation. 
In the semi-supervised setting, models are trained on the PASCAL-VOC(07+12) and COCO train sets without labels and are fine-tuned on $k\%$ of labeled data similar to~\cite{bar2021detreg}. In the fully supervised setting, pre-trained models are fine-tuned on the full PASCAL-VOC and COCO dataset using all the labels. For the class-agnostic object proposal generation, models are pre-trained on COCO dataset without labels and the generated object proposals are evaluated on the COCO validation set similar to~\cite{bar2021detreg}. 

We follow the settings used in~\cite{Vo21LOD} for multiple-object localization and evaluate on PASCAL-VOC 2007 and COCO20k. For single-object localization, we follow the settings in~\cite{LOST,wang2022tokencut} and evaluate on PASCAL-VOC 2007, PASCAL-VOC 2012 and COCO20k.

\noindent\textbf{\underline{Discovery setup}}:
This setup evaluates the object discovery performance. Similar to~\cite{LOST} we use the regions obtained by our localization method, to perform K-means clustering and use the resulting cluster labels to train Faster-RCNN object detectors on PASCAL-VOC 2007, 2012 trainval and COCO20k train sets. We report results of these experiments on the PASCAL-VOC 2007 test and COCO minival sets respectively. 
In addition to this, we report the performance of our discovery method on COCO train set, similar to the large scale discovery in~\cite{Rambhatla2021ThePO}.

\begin{figure*}[t]
 \begin{minipage}[t]{0.5\textwidth}
    \centering
    \includegraphics[width=\linewidth]{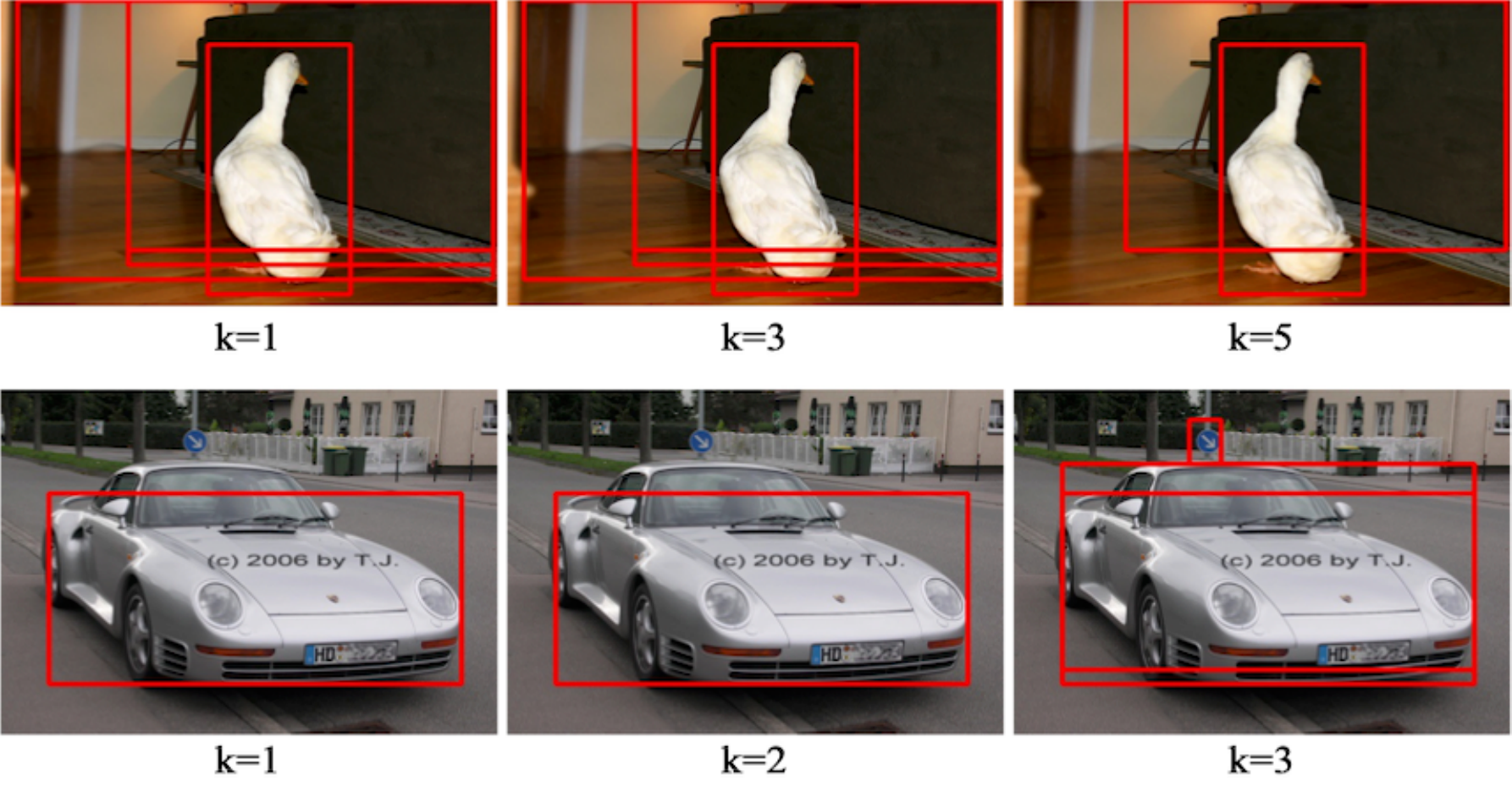}
    \captionof{figure}{\textbf{Effect of kernel size}: Different kernel sizes can identify different tokens as belonging to the foreground. Multiple kernels help eliminate noisy predictions (first triplet) and missed predictions (second triplet). }
    \label{fig:filt_size}
    \end{minipage}\quad
    \begin{minipage}[t]{0.5\textwidth}
    \centering
    \includegraphics[width=\linewidth]{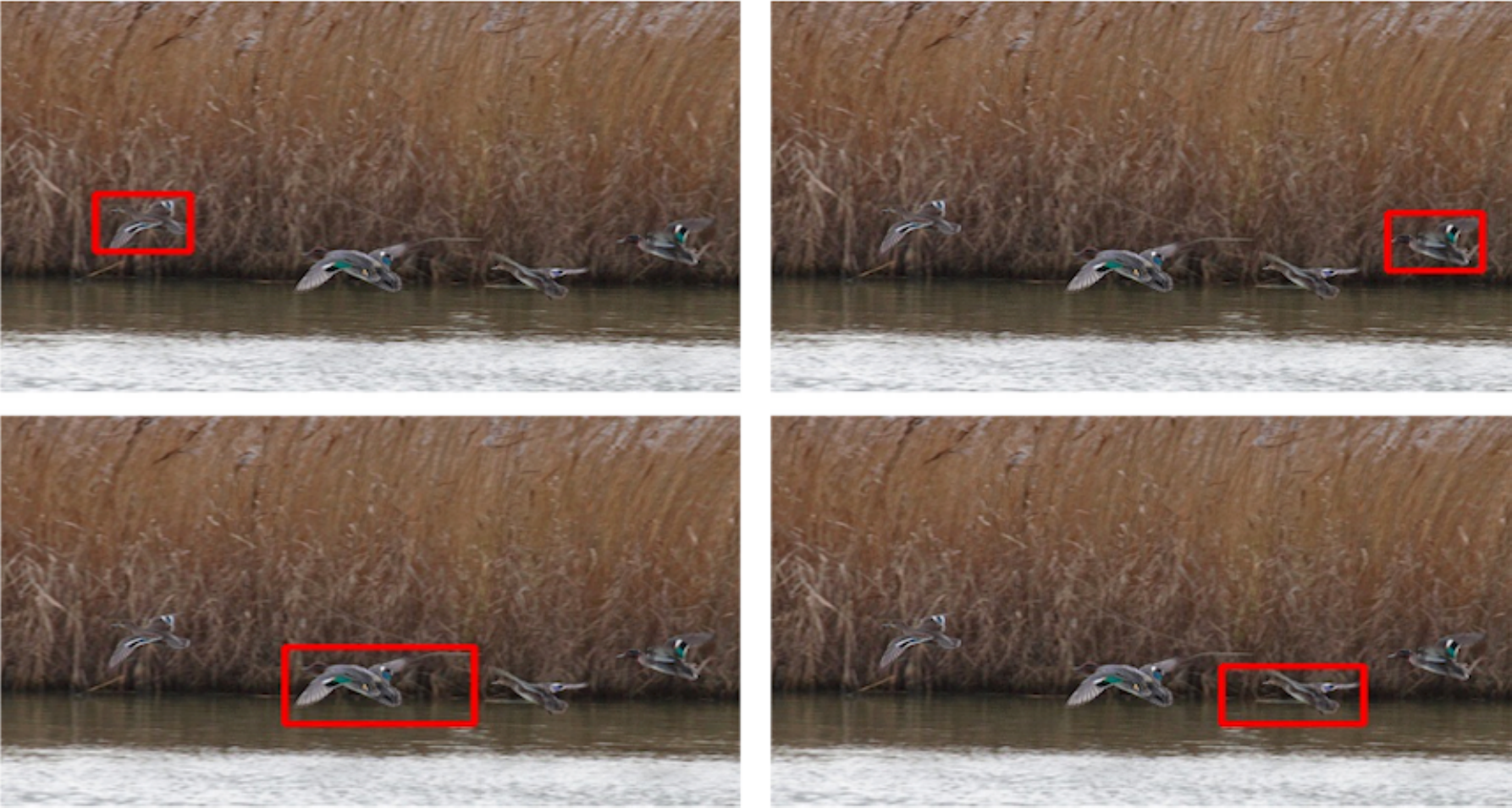}
    \caption{Figure demonstrating the effect of clustering in MOST: Each image consists of a bounding box generated from a \textit{pool}. We observe that each \textit{pool} focuses on different foreground instance. }
    \label{fig:clust_pool}
        \end{minipage}
\end{figure*}
\subsection{Comparison with contemporary methods}\label{subsec:results}
In this section we compare our method against contemporary works the \textit{localization} and \textit{discovery} setups.

\subsubsection{Localization setup}
\noindent\textbf{\underline{Unsupervised Pre-training}}: 
Table \ref{tab:od-1} compares the results of all the localization methods on unsupervised pre-training of object detectors. We use average precision at different IoU thresholds ([0.50:0.95]: AP, 0.5: AP$_{0.50}$, 0.75: AP$_{0.75}$) for evaluation.  On the semi-supervised setting, on VOC 07+12 ($k=10\%$), the self-supervised transformer based methods (LOST, TokenCut and MOST) outperform SS~\cite{Uijlings13} with fewer boxes per image. In particular, TokenCut (denoted as TCut in Table~\ref{tab:od-1}) which outputs only one box per image, outperforms SS, using ten boxes per image, by ${\sim}0.4$ points on mAP. MOST which outputs an average of $4.65$ boxes per image outperforms TokenCut (the best performing self-supervised transformer based method) by $1.89$, $2.7$ and $2.26$ percentage points on AP, AP$_{50}$, and AP$_{75}$ respectively. This can be attributed to the ability of MOST to output multiple foreground regions resulting in more samples for pre-training which is not possible in the case of TokenCut. MOST outperforms SS, that outputs 30 boxes per image, by $0.91$, $2.09$ and $0.9$ points on AP, AP$_{50}$, and AP$_{75}$ respectively using almost $6\times$ fewer boxes per image and this can be attributed to the ability of MOST to generate high quality proposals. On COCO, MOST outperforms TokenCut by $0.8$  and $0.86$ on the 1\% and 2\% setting of semi-supervised learning. 
MOST with a ViT-B/8 backbone, that outputs 13.09 boxes on average per image, outperforms SS (with 30 boxes per image) by $0.36$, $0.07$ points on k=1\% and k=2\% few shot splits of COCO respectively. 

On the fully supervised setting, MOST outperforms LOST and TokenCut by 0.76 and 0.55 (AP) percentage points respectively on VOC 07+12. On COCO, MOST outperforms them by 0.50 and 1 points respectively.
On VOC 07+12, MOST using ViT-B/8 (13.09 boxes per image) outperforms SS (with 30 boxes per image) by 0.40. On the much harder COCO dataset, MOST outperforms SS 1.20 (AP) percentage points using $2\times$ fewer boxes per image.

In Table~\ref{tab:caop} we report the class agnostic object proposal evaluation of DETReg trained using different localization methods. We report average precision at different IoU thresholds (AP, AP$_{50}$, AP$_{75}$) and recall $@$ 1, 10 and 100 proposals per image (denoted as R1, R10, and R100) to evaluate the quality of region proposals. Note that the numbers in the table are low because of the unsupervised nature of training. All the self-supervised transformer-based methods achieve performance better than SS with far fewer boxes. In recall, TokenCut and MOST perform on par with each other and outperform rest of the methods with significant improvements. MOST achieves the highest performance on average precision among all the methods. It can achieve higher precision and recall because of its ability to output multiple high quality regions per image. While LOST and TokenCut output high quality boxes, they cannot output more than one box per image. SS on the other hand, outputs multiple boxes but with poor quality.  

\noindent\textbf{\underline{Multiple Object Localization}}:
We compare with LOD, the state-of-the-art method on the multi-object localization benchmark proposed by LOST using the code released by authors. On VOC2007, we attain an odAP[0.5:0.95] of 6.43 compared to 5.35 attained by LOD, an improvement of 1.09 percentage points. On the COCO20k dataset, we attain a performance of 1.70 (compared to 1.53 achieved by LOD) on the harder odAP[0.50:0.95] metric. Note, we do not compare with rOSD~\cite{Vo20rOSD} as LOD~\cite{Vo21LOD} outperforms it. 

\noindent\textbf{\underline{Single Object Localization}}: 
Table \ref{tab:corloc} compares the results of our method on single object localization with recent methods on PASCAL VOC 2007, 2012 and COCO20k respectively. We use the CorLoc metric to evaluate methods. Note that MOST is a multiple object localization method and this setup evaluates the ability of methods to output a single region. Since MOST outputs multiple boxes, we use the heuristic, average best overlap (for evaluating object proposals in \cite{Uijlings13}), to select one region per image. The numbers reported for MOST in this table are the ``best" case scenario.  We outperform LOST by $12.9$, $13.4$ and $16.4$ percentage points on VOC 2007, 2012 and COCO20k respectively. We outperform TokenCut~\cite{wang2022tokencut} by 6, 5.3 and 8.3 percentage points on the three datasets respectively.
To obtain multiple regions per image, authors of LOST train a foreground object detector using the regions obtained by their method as supervision, called LOST+CAD~\cite{LOST}. This method can output multiple boxes per image and from Table \ref{tab:corloc}, even without any training, our method outperforms LOST+CAD and TokenCut+CAD by 9.1, 7, 9.6 and 3.4, 2.1, 4.5  percentage points on VOC 2007, 2012 and COCO20k respectively.

\noindent\textbf{\underline{Discovery Setup}}:
This setup evaluates the true object discovery performance as the localized boxes are used to discover semantic groups.
Following LOST~\cite{LOST}, we first cluster the features of the localized objects using K-means clustering. 
For VOC 2007 and 2007+2012 trainval splits, we use 20, 30 and 40 clusters. We use 80, 90 and 100 clusters for COCO20k train split. We report the results of experiments on VOC 2007, 07+12 trainval sets on VOC 2007 test set. For experiments on COCO20k, we report results on the COCO validation set. Results are tabulated in Table \ref{tab:od1}. MOST outperforms LOST in most settings with the margin of improvement higher for more number of clusters and more cluttered datasets like COCO. For more details on clustering and training refer to supplementary.
\begin{figure}
    \centering
    \includegraphics[width=\linewidth]{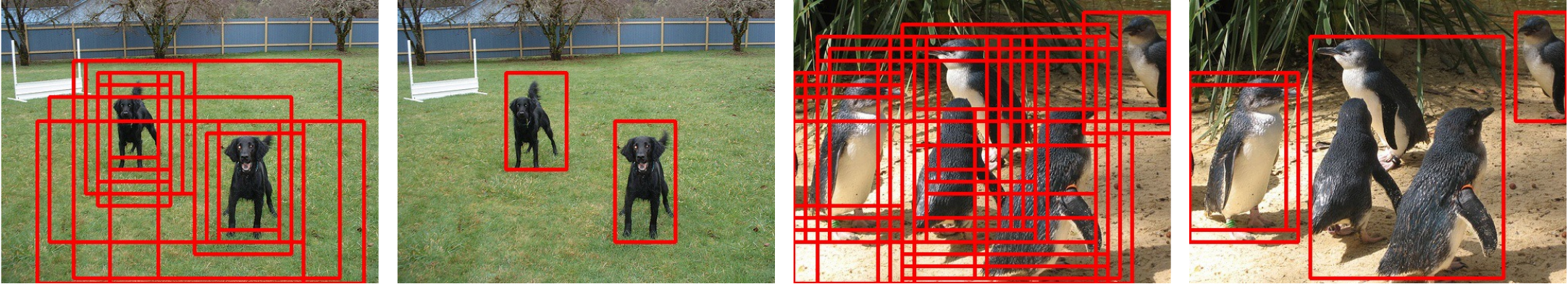}
    \caption{Figure demonstrating the effect of clustering in MOST. Eliminating the clustering results in noisier outputs. }
    \label{fig:clust}
\end{figure}

Finally, we evaluate the performance of MOST on large-scale object discovery setup introduced in~\cite{Rambhatla2021ThePO}.
For this setup, we use the area under the purity coverage plot as the metric. \cite{Rambhatla2021ThePO} automatically identifies the number of clusters and obtains an AuC@0.5 of 3.6\% on the COCO 2014 train set.
We extract the DINO [\texttt{CLS}] token features of regions obtained from MOST for K-Means clustering. To avoid specifying the number of clusters manually, we employ the ``kneedle" method~\cite{5961514} to get the optimal number of clusters (more details in supplementary).
Next, we randomly sample 10000 features from the whole dataset and cluster them using K-means with the optimal number of clusters. This subsampling avoids loading all the features into memory. MOST + optimal K-means achieves an AuC@0.5 of 8.74\% on COCO 2014 train set. 
We use the cluster labels to train an object detector on the COCO train set and achieve an AP/AP$_{50}$ of 3.9/9.5\% compared to  5.2\% AP$_{50}$ obtained by~\cite{Rambhatla2021ThePO} on COCO validation set. For more experiments on unsupervised saliency detection and weakly supervised localization, refer to the supplementary.

\begin{figure}[t]
    \centering
    \includegraphics[width=\linewidth]{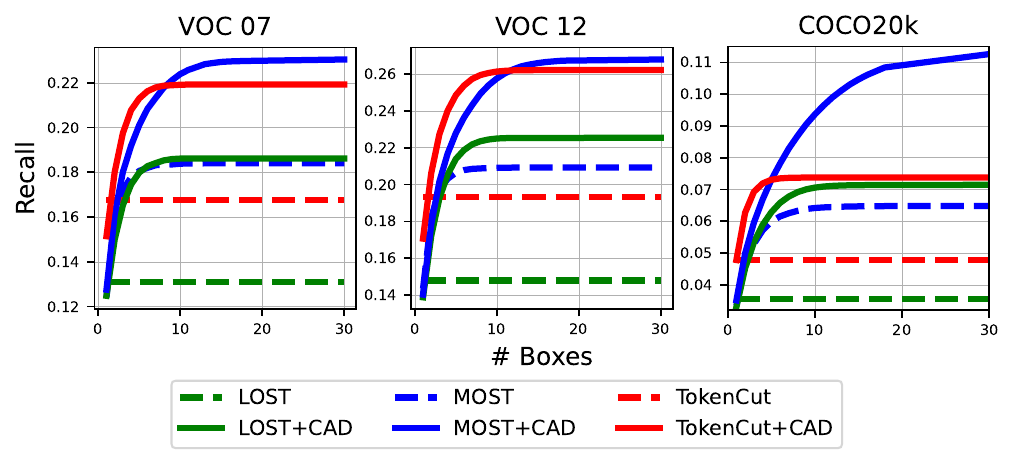}
    \caption{\textbf{Recall analysis}: Comparison of recall values of MOST, MOST+CAD with LOST and LOST+CAD. LOST generates one bounding box per image. MOST+CAD, MOST have higher recall and cover more ground-truth objects for a fixed set of boxes.}
    \label{fig:recall}
\end{figure}

\subsection{Ablation Experiments}\label{subsec:ablation}
\noindent\textbf{\underline{Effect of kernel size}}: The EBA module performs box analysis in a sliding window fashion using boxes (or kernels) of different sizes. 
We implement this efficiently using a pooling operation. 
We visualize the effect of the size of pooling kernels on the final output in Fig.~\ref{fig:filt_size}. We observe that the majority voting performed in EBA, helps in removing noisy predictions in the first triplet, where a box identified by kernel of size 1 is eliminated by majority voting of kernels with larger receptive field. In the second triplet in Fig.~\ref{fig:filt_size}, an object which was missed by the lower order kernels (k=[1,4]), can be picked up with a higher order kernel (k=5). 
For all our experiments, we use kernels of sizes upto 5 (\ie. k=[1,5]). For quantitative results on the effect of kernel sizes, refer to the supplementary.

\noindent{\textbf{\underline{Effect of clustering}}}:
MOST performs clustering with the token locations as features to obtain \textit{pools}. 
Each pool contains tokens belonging to a foreground object. We show the effect of clustering qualitatively in Fig.~\ref{fig:clust_pool}. 
We observe that each pool focuses on one foreground object and illustrate the bounding boxes extracted from each \textit{pool}.

Each pool contains tokens belonging to a foreground object. We show the effect of forming \textit{pools} qualitatively in Fig. \ref{fig:clust}. The first image in each pair shows the output of MOST without clustering to form \textit{pools} and the second image shows the output with the formation. We observe that, without clustering, each token can generate a bounding box resulting in noisier outputs.

\noindent\textbf{\underline{Recall of boxes}}:
To analyze the object localization performance of MOST, we compare its recall with LOST and TokenCut on VOC 07, 12 and COCO20k datasets in Fig.~\ref{fig:recall}. 
The x-axis represents the maximum number of boxes allowed per image and the y axis plots the recall. LOST and TokenCut generate only one box per image and hence have fixed recall in all the plots. MOST can generate more boxes and hence have higher recall than LOST and TokenCut. \cite{LOST} trains a class agnostic detector (CAD) to output multiple boxes per image using the output of LOST as supervision. Without a single step of training, MOST performs competitively against LOST+CAD on all the datasets. With a class agnostic detector, MOST+CAD outperforms LOST, TokenCut and their CAD counterparts comfortably on all the datasets.
On COCO20k, a much harder dataset, MOST+CAD outperforms all the methods with a significant margin demonstrating its superior localization abilities.

\noindent\textbf{\underline{Effect of EBA}}: We study the effect of EBA on single-object localization. The task of the EBA module is to identify tokens on foreground instances from similarity maps. We replace the EBA module with the strategy used by LOST~\cite{LOST}, effectively giving LOST the ability to localize multiple objects. We use top-100 patches and this system achieves a CorLoc of 63.66 (compared to 74.84 of MOST). The EBA module can automatically pick the right tokens, unlike LOST to localize multiple objects. This experiment demonstrates the benefit of the proposed EBA module.
\begin{figure*}[t]
    \centering
    \begin{subfigure}[t]{0.49\linewidth}
        \centering
        \includegraphics[width=\linewidth]{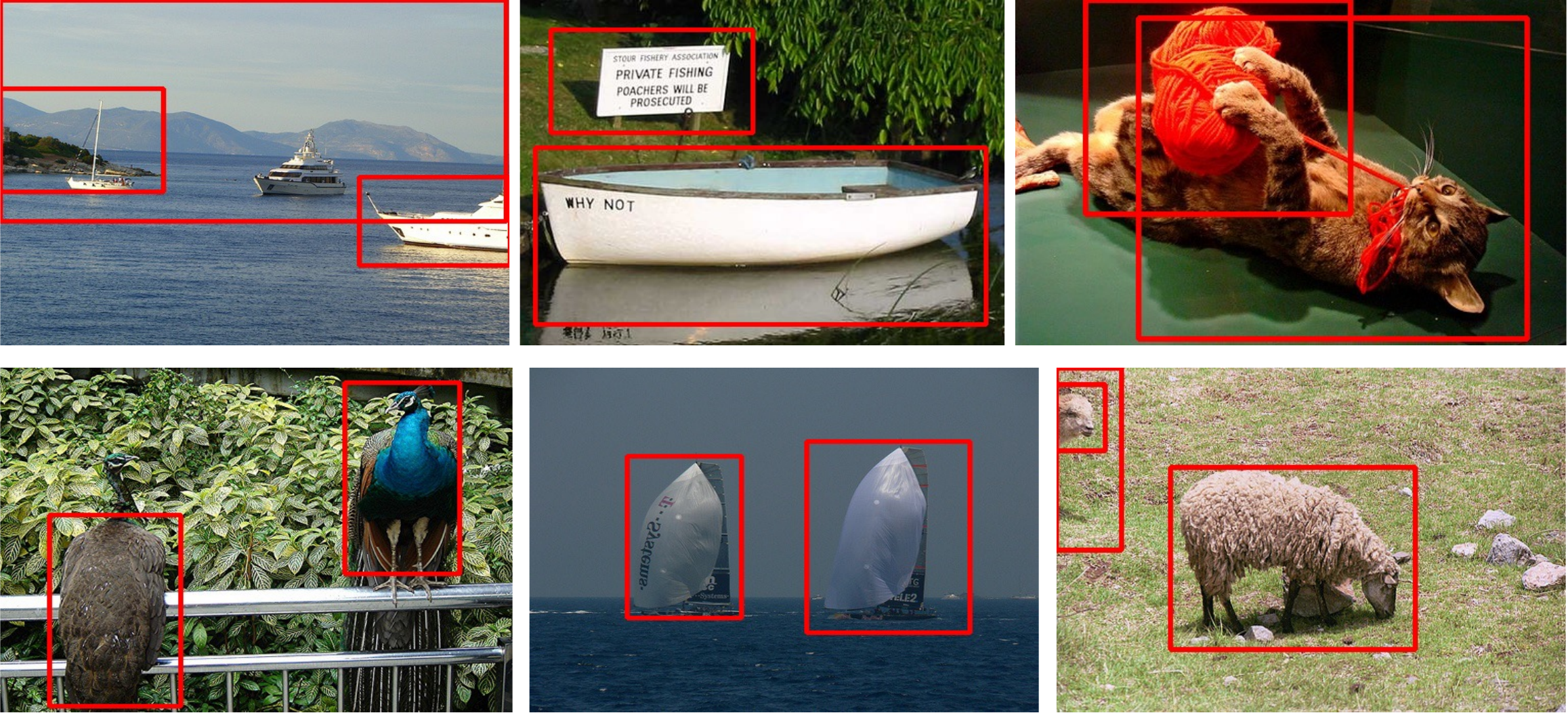}
        \caption{VOC 07+12}\label{fig:quant_viz_voc07}
    \end{subfigure}
    \begin{subfigure}[t]{0.49\linewidth}
        \centering
       \includegraphics[width=\linewidth]{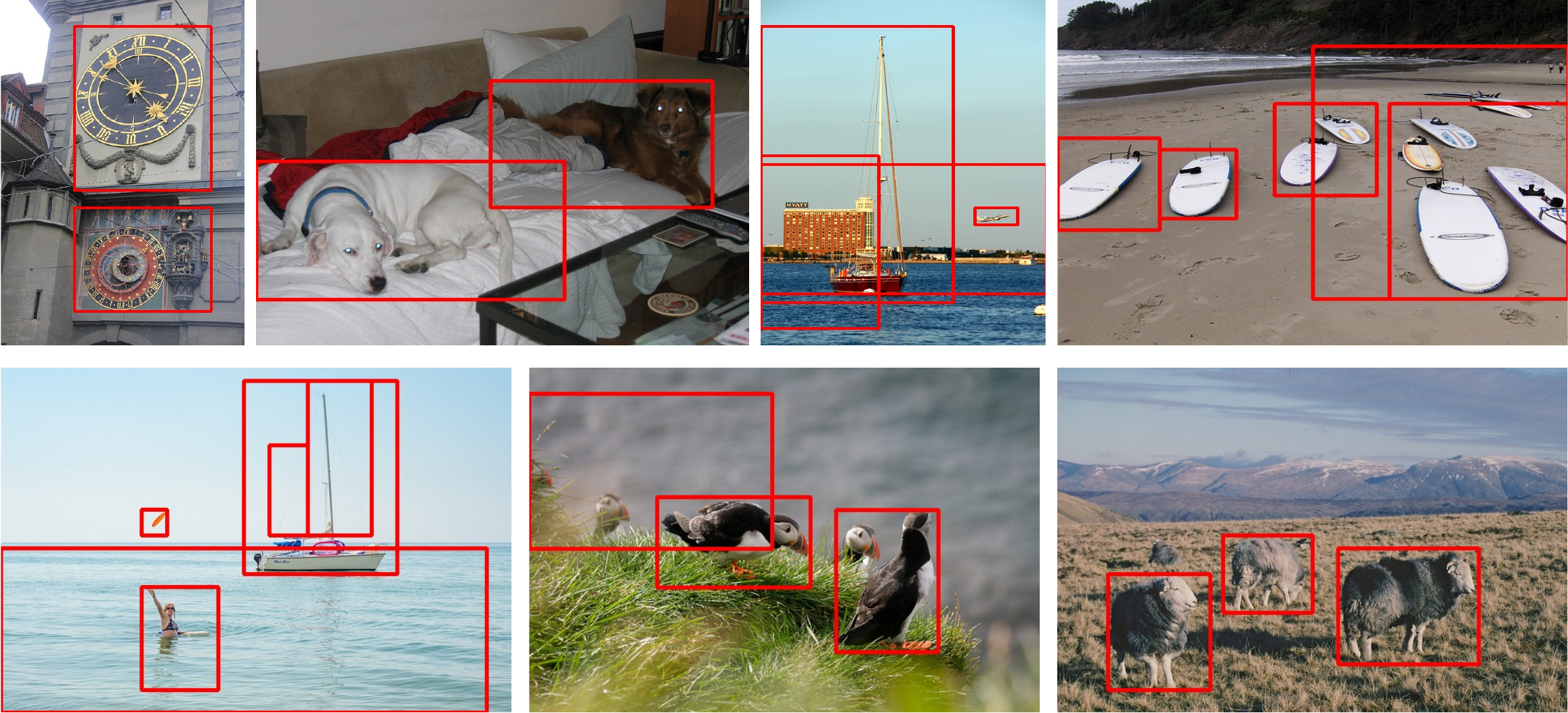}
        \caption{COCO}\label{fig:quant_viz_coco}
    \end{subfigure}
    \caption{\textbf{Qualitative results of MOST on VOC 07, 12 and COCO}: MOST can localize multiple objects in cluttered scenes without training. Localization ability of MOST is not limited by the biases of annotators and can localize rocks, mountains, branches, water bodies, \etc.}
    \label{fig:quant_viz}\vspace{-1.5em}
\end{figure*}

\noindent{\textbf{\underline{Timing analysis}}}:
MOST can localize multiple objects per image and does so by analyzing the similarity maps of all the tokens which is computationally more expensive. To understand this requirement, we perform a timing analysis. On average, LOST takes 0.008s per image while MOST takes 0.3s. MOST obtains a recall of 0.19, 0.21, and 0.08 on VOC07, VOC12, and COCO20k respectively (compared to 0.13, 0.15, and 0.03 of LOST). This translates to localizing an additional 750, 2400, and 7200 instances than LOST on VOC07, VOC12, and COCO20k respectively. We believe the additional time taken by MOST is justified by the improvement in recall performance which is essential for an object localization method.
\begin{table}
\setlength{\tabcolsep}{3pt}
\renewcommand{\arraystretch}{1.2}
\centering
\footnotesize
\caption{Effect of backbone on single-object localization }\label{tab:corloc_bbone}
\begin{tabular}{@{}lccc@{}}
\toprule
 \textbf{Backbone}& \textbf{VOC 2007} & \textbf{VOC 2012}& \textbf{COCO20k}  \\
 \midrule
 ViT-S/8 & 84.25& 86.00 & 80.50\\
 ViT-S/16 & 74.80 & 77.40 & 68.60\\
 ViT-B/8 & 85.40 & 87.00& 81.73\\
 ViT-B/16 & 72.72 & 76.28 & 67.20\\
\bottomrule
\end{tabular}
\end{table}

\noindent\textbf{\underline{Effect of backbone}}:
We study the effect of backbone on MOST in Table \ref{tab:corloc_bbone}. We observe that backbones with a smaller patch size can localize more objects, especially smaller ones (refer to supplementary for qualitative results) resulting in a higher CorLoc. This comes at a cost of producing noisier outputs. We refer interested readers to the supplementary for more analysis and qualitative results on the effect of backbones and patch sizes. 
\subsection{Additional Experiments}
\noindent\textbf{\underline{Unsupervised Saliency Detection}}:
MOST can easily be extended to unsupervised saliency detection. We experiment with ECSSD~\cite{Shi2016HierarchicalIS}, DUTS~\cite{wang2017} and DUT-OMRON~\cite{yang2013saliency} datasets and use the metrics used by~\cite{LOST,wang2022tokencut} for evaluation. All these datasets require methods to segment the salient object in an image. Hence, naively using MOST doesn't perform well. To extend MOST for saliency detection, we select the largest \textit{pool} and use the similarity map computed using its tokens as the saliency map. In Table \ref{tab:saliency}, we compare MOST with TokenCut and LOST on the three datasets. MOST outperforms LOST comfortably on all the metrics, datasets and fares competitively against TokenCut. We believe that the ability to detect multiple objects in images is a good tradeoff for a slight drop in performance on saliency detection. We refer interested readers to the supplementary for qualitative results on saliency detection.

\begin{table}
\begin{minipage}[b]{\linewidth}
\setlength{\tabcolsep}{1pt}
\renewcommand{\arraystretch}{1.4}
\centering
\footnotesize
\caption{\textbf{Results on unsupervised saliency detection}: We compare MOST to state-of-the-art unsupervised saliency detections methods. Top results are in bold and second best is highlighted in \textcolor{blue}{blue}. Ability to detect multiple objects in images is a good tradeoff
for a slight drop in performance on saliency detection}\label{tab:saliency}
\resizebox{\linewidth}{!}{
\begin{tabular}{@{}lccccccccc@{}}
\toprule
 \textbf{Method}& \multicolumn{3}{c}{\textbf{ECSSD}} & \multicolumn{3}{c}{\textbf{DUTS}}& \multicolumn{3}{c}{\textbf{DUT-OMRON}}  \\
  \cmidrule[\cmidrulewidth](l){2-4}
 \cmidrule[\cmidrulewidth](l){5-7}
 \cmidrule[\cmidrulewidth](l){8-10}
  & max $F_{\beta}$ & IoU & Acc (\%) & max $F_{\beta}$ & IoU & Acc (\%) & max $F_{\beta}$ & IoU & Acc (\%) \\
 \midrule
 DeepUSPS\cite{Nguyen2019DeepUSPSDR} & 58.4 & 44.0 & 79.5 & 42.5 & 30.5 & 77.3 & 41.4 & 30.5 & 77.9\\
 BigBiGAN~\cite{Voynov2021ObjectSW} & 78.2 & 67.2 & 89.9 & 60.8 & 49.8 & 87.8 & 54.9 & 45.3 & 85.6\\
 E-BigBiGAN~\cite{Voynov2021ObjectSW}& \textcolor{blue}{79.7} & \textcolor{blue}{68.4} & \textcolor{blue}{90.6} & 62.4 & 51.1 & 88.2 & 56.3 & 46.4 & 86.0\\
LOST~\cite{LOST} & 75.7 & 62.5 & 88.0& 61.8 & 52.0 & 87.1 & 47.4 & 40.2 & 79.6\\

 TCut~\cite{wang2022tokencut} & \textbf{80.3} & \textbf{71.2} & \textbf{91.8} & \textbf{67.2} & \textbf{57.6} & \textbf{90.3} & \textbf{60.0} & \textbf{53.3} & \textbf{88.0}\\
 MOST & 79.1 & 63.1 & 89.0 & \textcolor{blue}{66.6} & \textcolor{blue}{53.8} & \textcolor{blue}{89.7} &\textcolor{blue}{57.0} & \textcolor{blue}{47.5} & \textcolor{blue}{87.0}\\
\bottomrule
\end{tabular}
}
\end{minipage}\vspace{-1.5em}
\end{table}

\noindent\textbf{\underline{Weakly supervised localization:}}
We evaluate MOST on weakly supervised object localization on CUB-200~\cite{399} and Imagenet~\cite{ILSVRC15} datasets respectively, and achieve a CorLoc of 92.42 (\vs 91.8 of TokenCut) on CUB-200 and a CorLoc of 71.4 (\vs 65.4 of TokenCut).

\subsection{Qualitative Results:}\label{subsec:q_res}
We illustrate qualitative results of MOST on VOC2007, 2012 and COCO datasets in Fig.~\ref{fig:quant_viz}. Fig.~\ref{fig:quant_viz_voc07} shows results on VOC 2007 and 2012 and Fig.~\ref{fig:quant_viz_coco} shows results on COCO dataset. MOST is capable of localizing fairly complex scenes in all the three datasets. We observe that, such unsupervised localization methods are not limited by the categories annotated by humans but can localize regions  of ``stuff" like sign boards (third image in the first row of Fig.~\ref{fig:quant_viz_voc07} right), rocks (last image of last row in Fig.~\ref{fig:quant_viz_coco}), and water bodies (first image in second row of Fig.~\ref{fig:quant_viz_coco}).

\section{Conclusion}\label{sec:conclusion}
We present MOST, an effective method for localizing multiple objects in complex images without a single annotation. MOST leverages object segmentation properties of transformers trained using DINO~\cite{caron2021emerging}. We show that the ability of MOST to localize multiple objects in an image is very effective on several object localization and discovery benchmarks. In particular, MOST outperforms recent state-of-the-art methods that train a class agnostic detector, on the task of single object localization, without any training. Further, we show that MOST achieves higher recall and covers more ground truth objects for a fixed set of boxes than LOST~\cite{LOST}, a contemporary work on object localization. Finally, we extend MOST to the task of unsupervised saliency detection and report competitive results with recent works.

\smallskip
\noindent\textbf{Acknowledgement.} This project was partially supported by DARPA SemaFor (HR001119S0085) and DARPA SAIL-ON (W911NF2020009), and Amazon Research Award to Abhinav Shrivastava. Rama Chellappa was supported by an ONR MURI (N00014-20-1-2787).

{\small
\bibliographystyle{unsrtnat}
\bibliography{egbib}
}

\end{document}